\documentclass[10pt,twocolumn,letterpaper]{article}

\usepackage{cvpr}              
\usepackage{graphicx}
\usepackage{color,xcolor,colortbl}
\definecolor{lightgray}{gray}{0.95}
\definecolor{lightblue}{HTML}{95bddc}
\usepackage{bbding}
\usepackage{adjustbox}
\usepackage{makecell}
\usepackage{multirow} 
\usepackage{tcolorbox}
\usepackage[accsupp]{axessibility}
\definecolor{cvprblue}{rgb}{0.21,0.49,0.74}
\usepackage[pagebackref,breaklinks,colorlinks,allcolors=cvprblue]{hyperref}

\def\paperID{4014} 
\def\confName{CVPR}
\def\confYear{2025}

\title{DreamOmni: Unified Image Generation and Editing}

\author{	Bin Xia $^{1}$, Yuechen Zhang  $^{1}$, Jingyao Li$^{1}$, Chengyao Wang $^{1}$, \\Yitong Wang$^{2}$, Xinglong Wu$^{2}$, Bei Yu$^{1}$, and Jiaya Jia $^3$ \\
	$^{1}$ CUHK, $^2$  ByteDance Inc, $^3$ HKUST  \\
    \href{https://zj-binxia.github.io/DreamOmni-ProjectPage/}{https://zj-binxia.github.io/DreamOmni-ProjectPage/}
}

\begin{document}

\twocolumn[{
\maketitle
\begin{center}
    \vspace{-8mm}
    \captionsetup{type=figure}
    \includegraphics[width=1.0\linewidth]{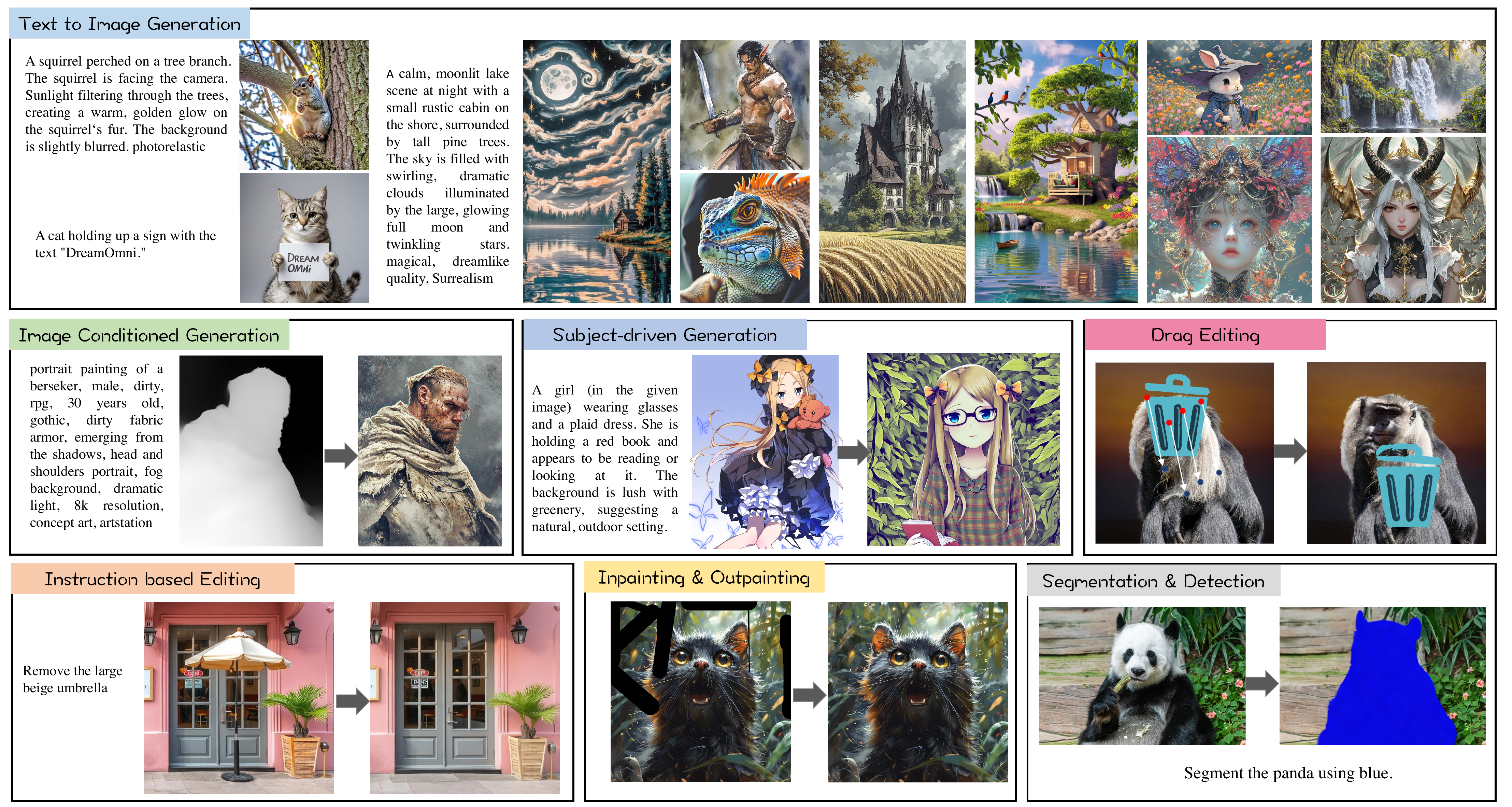}
    \vspace{-5mm}
    \captionof{figure}{The gallery of DreamOmni. DreamOmni, as a native unified image generation and editing model, can handle various tasks. }
    \label{fig:summary}
    \vspace{1mm}
\end{center}
}]

\begin{abstract}
\vspace{-3mm}
Currently, the success of large language models (LLMs) illustrates that a unified multitasking approach can significantly enhance model usability, streamline deployment, and foster synergistic benefits across different tasks. However, in computer vision, while text-to-image (T2I) models have significantly improved generation quality through scaling up, their framework design did not initially consider how to unify with downstream tasks, such as various types of editing. To address this, we introduce DreamOmni, a unified model for image generation and editing. We begin by analyzing existing frameworks and the requirements of downstream tasks, proposing a unified framework that integrates both T2I models and various editing tasks. Furthermore, another key challenge is the efficient creation of high-quality editing data, particularly for instruction-based and drag-based editing. 
To this end, we develop a synthetic data pipeline using sticker-like elements to synthesize accurate, high-quality datasets efficiently, which enables editing data scaling up for unified model training.  For training, DreamOmni jointly trains T2I generation and downstream tasks. T2I training enhances the model's understanding of specific concepts and improves generation quality, while editing training helps the model grasp the nuances of the editing task. This collaboration significantly boosts editing performance. Extensive experiments confirm the effectiveness of DreamOmni. The code and model will be released. 
\end{abstract}    
\vspace{-6mm}
\section{Introudction}
Recently, text-to-image (T2I) generative foundation models~\cite{LDM,sdxl,sd3,imagen,dalle3,playground25} have made remarkable progress, driving the development of various downstream applications such as image editing, video generation, and more.

Currently, T2I foundation models face two significant challenges: \textbf{(1)} Adapting these models for downstream applications often requires the integration of various plugins (such as ControlNet~\cite{controlnet} and IP-adapter~\cite{ip-adapter}) in different ways, or the extension of input channels (\eg, SD-inpainting~\cite{LDM}, InstructP2P~\cite{instructpix2pix}). This reliance on specialized frameworks hampers multi-task generalization and complicates deployment. \textbf{(2)} High-quality and accurate editing data is difficult to obtain, including instruction-based editing, drag editing, and subject-driven generation data. In this paper, we propose to unify T2I models with multiple editing tasks, such as instruction-based editing, inpainting \& outpainting, drag editing, and reference image generation within a single framework. Additionally, we introduce an efficient synthetic data pipeline for efficiently and accurately constructing editing data, facilitating the training of native unified generation and editing model.

We began by exploring fundamental model frameworks to develop a unified framework that is both highly effective and efficient, achieving rapid training convergence. \textbf{(1)} We aligned certain frameworks, such as Unet~\cite{LDM} and DIT~\cite{DIT}, to have similar parameter sizes and runtime settings for fair comparison (Fig.~\ref{fig:combined_comparison}). Our analysis showed that DIT’s effectiveness comes from concentrating most of its computational and parameter load on latent at 2$\times$ downsampling size, whereas Unet distributes more of its operations at smaller scales. This allocation makes DIT comparatively superior, as concentrating computations at the 2$\times$ downsampled latent achieves a better trade-off. However, we observed that Unet’s residual connections significantly improve the model's training convergence speed. Inspired by the efficiency of high-resolution latent computations, we also incorporated additional residual convolutional networks for the input-scale latent. \textbf{(2)}  We replaced the original text encoder with a vision-language model (VLM) to unify visual-language prompt encoding. This encoded prompt was then concatenated with the noisy latent, allowing for integrated computation within the DIT framework.

\begin{figure*}[t]
	\centering
 \resizebox{1\linewidth}{!}{
	\includegraphics[height=8cm]{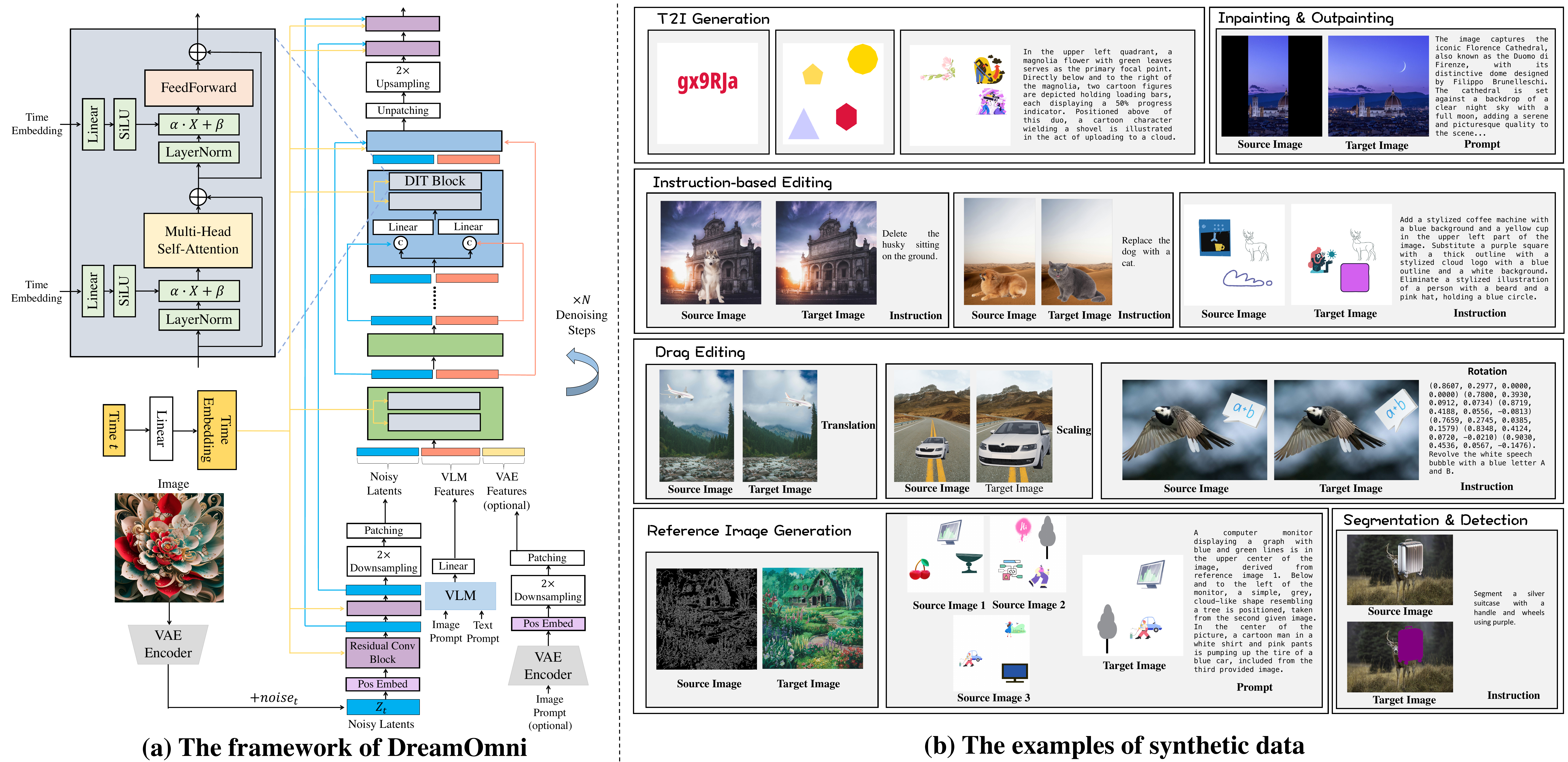}
 }
\vspace{-6mm}
	\caption{ 
 The overview of DreamOmni.
\textbf{(a)} The DreamOmni framework supports unified image generation and editing, with fast training convergence and powerful performance.
\textbf{(b)} To overcome the difficulty and inefficiency in data creation and filtering for image editing, we propose a collage-based synthetic data pipeline. This pipeline enables the efficient creation of data for various editing tasks, such as adding, deleting, and replacement operations in instruction-based editing, as well as translation, scaling, and rotation in drag editing. Additionally, it supports reference image generation and segmentation $\&$ detection. Furthermore, our synthetic data pipeline enhances the accuracy of T2I generation.
Due to space limitations, we have optionally shown the corresponding prompts or instructions for these cases. }
\vspace{-4mm}
\label{fig:overall}

\end{figure*}

Next, we need to obtain accurate training data efficiently for our unified model. For various types of editing, such as instruction-based editing, it’s challenging to create and filter pairs of data that precisely match the instructions while avoiding detail corruption. We have discovered that the key to effective editing lies in helping the model understand the meaning of the editing operations, rather than learning specific concepts (a capability the model already possesses in T2I training). Therefore, we propose a synthetic collage data pipeline that easily scales up to generate accurate and diverse edited data, enabling models to better understand the nuances of specific editing tasks (Fig.~\ref{fig:overall} (b)).
\textbf{(1)} For instruction-based editing,  we constructed synthetic data for three main categories: addition, removal, and replacement.
\textbf{(2)} For drag editing, we utilized synthetic data to create scaling, translating, and rotating editing data.
\textbf{(3)} For inpainting and outpainting, we randomly generate masks for each given image.
\textbf{(4)} For reference image generation, We divided the tasks into subject-driven generation and image-conditioned generation (\ie, ControlNet-like generation) tasks. For subject-driven generation, we used a canvas filled with stickers, allowing the model to reference a specific sticker for generation, which helps train the model to extract the demanded objects or details from given images for generation. For the image-conditioned generation, we created canny, depth, and segment maps, similar to ControlNet. \textbf{(5)} We use synthetic data to enhance the model's T2I accuracy in responding to attributes such as quantity, position, relationships, color, shape, and text. Concretely, we randomly arrange different quantities of stickers, text, and geometric shapes on a canvas, and obtain precise descriptions of their quantity, position, relationships, and colors based on their exact coordinates.

During training, thanks to our unified framework that integrates T2I with multiple editing tasks, we can easily train T2I data alongside various types of editing data, obtaining a native unified image generation and editing model, called DreamOmni. The editing tasks benefit from the T2I data, which helps prevent the model from forgetting specific concepts and generation quality decline, while the T2I task can leverage synthetic data to enhance instruction following. 

\begin{itemize}

\item We conducted an analysis of existing model frameworks under fair settings and, drawing on the characteristics of different tasks, proposed an efficient and powerful unified image generation and editing framework, DreamOmni.

\item We introduced a synthetic collage data pipeline to tackle the current inefficiency and difficulty in creating and filtering high-quality editing data. Furthermore, we utilized the synthetic collage data pipeline to enhance the accuracy of the T2I model’s output. Experimental results demonstrate that synthetic data is an effective, high-quality, and cost-efficient method to scale up data for achieving unified image generation and editing training.

\item After unified training with T2I data and various synthetic datasets, our DreamOmni showcases competitive performance on T2I generation and various editing tasks.

\end{itemize}

\vspace{-1mm}
\section{Related Work}

\noindent\textbf{Text-to-image Diffusion Model.} 
Diffusion models~\cite{sohl2015deep,DDPM,DDPM3,DDPM4,DDPM5,DDPM6,batzolis2021conditional,song2020score,glide} have emerged as highly effective methods for image generation, outperforming previous generative models like GANs~\cite{GAN} and VAEs~\cite{VAE}. The latent diffusion model~\cite{LDM}, also known as Stable Diffusion, enhances the diffusion process within the image latent space and utilizes cross-attention to guide image generation through text input. This approach enhances the usability of the image generation model, driving its widespread adoption. Subsequently, more advanced text-to-image (T2I) models were developed, such as Imagen~\cite{imagen}, DALLE~\cite{dalle}, and others~\cite{dalle3,ediff,ernie-vilg2,sdxl,dai2023emu}. 
Recently, the DIT frameworks~\cite{DIT,UVIT} has become increasingly popular as more models~\cite{sd3, pixart, hunyuan} transition from the Unet framework~\cite{unet}, thanks to its straightforward structure and outstanding compatibility with multimodal inputs. However, the current T2I model fails to account for the design of downstream editing tasks, necessitating numerous specialized adaptations. For instance, Instructp2p~\cite{instructpix2pix} requires an expansion of input channels, while ControlNet~\cite{controlnet} and IP-adapter~\cite{ip-adapter} rely on additional plugins and various methods for injecting reference image information. These fragmented designs are inconsistent with the trend toward unified models, complicating deployment and impeding the joint training of T2I and editing tasks. In this paper, we will analyze the existing T2I frameworks and propose a more efficient and powerful unified generation and editing model.

\noindent\textbf{Multimodal-guided Image Editing and Generation} offers richer guiding information and has broader applications~\cite{edict,ledits++,p2pzero,masactrl,pnp,blendeddiff,diffedit} than T2I models. It encompasses instruction-based editing~\cite{instructpix2pix,magicbrush,emu-edit,smartedit,instructdiffusion}, inpainting \& outpainting~\cite{manukyan2023hd,llmga,smartbrush}, drag editing~\cite{dragdiffusion,freedrag}, and reference image generation~\cite{controlnet,blipdiffusion,controlnetpp,unicontrol,Uni-ControlNet}.
A significant challenge in editing, especially in instruction-based and drag editing, lies in creating and filtering accurately paired editing data. For example, InstructP2P~\cite{instructpix2pix} utilizes finetuned GPT-3 and a Prompt-to-Prompt strategy~\cite{p2p} to generate instruction-based editing datasets, but it achieves a success rate of under $15\%$ with frequent artifacts and detail loss. Additionally, filtering high-quality editing data is a persistent challenge, complicating scaling up.
Magicbrush~\cite{magicbrush} addresses this by employing people to create and filter editing data, but this manual method limits data volume. Furthermore, for drag editing,  most methods~\cite{dragdiffusion,freedrag} are training-free, which restricts their performance and efficiency. To this end, InstaDrag~\cite{instadrag} leverages video data to construct drag editing data. However, InstaDrag still faces challenges with inefficiency in filtering valid data from video. Moreover, for subject-driven generation~\cite{blipdiffusion,Kosmos-g}, DreamBooth~\cite{ruiz2023dreambooth} requires training on several subject-related images before each inference, which is inconvenient and inefficient.
Some works~\cite{li2024photomaker,xiao2024omnigen} collect multiple photos of the same person from the internet to build datasets for training more efficient and powerful subject-driven generation models. However, the process of gathering and filtering such data still presents a challenge, limiting the scaling up of the dataset. In this paper, we propose a simple, efficient, and accurate synthetic collage data pipeline to create quite a few data for unified training. 
\vspace{-1mm}
\section{Methodology}

The unification of multiple tasks is a trend and pursuit in the field of computer vision and AI, which not only enhances model usability and reduces deployment complexity but also enables collaborative training that fosters synergies between tasks. However, current T2I foundation models are primarily designed specifically for T2I and often overlook the potential for integration with other tasks, such as various image editing tasks. To this end, we propose DreamOmni, a unified model for image generation and editing. We design and train DreamOmni from three aspects. \textbf{(1)} In Sec.~\ref{sec:framework}, we compare various frameworks in a fair setting and design a powerful and fast training converging framework that supports unified multi-tasking based on the characteristics of different tasks. \textbf{(2)} In Sec.~\ref{sec:data}, we aim to propose a convenient, efficient, and accurate synthetic data pipeline for data scaling up to facilitate multi-task unified training and enhance the model's instruction-following ability. \textbf{(3)} In Sec.~\ref{sec:training}, we introduce the training scheme for DreamOmni. Native unified training of T2I and various editing tasks prevents concept forgetting and generation quality decline, while enhancing the model's editing and prompt-following capabilities.

\subsection{Framework}
\label{sec:framework}
In this part, we aim to design a unified and powerful image generation and editing framework.
Currently, different editing models often have distinct structure designs. For example, IP-adapter~\cite{ip-adapter} and BLIP-Diffusion~\cite{blipdiffusion} inject information through cross-attention to maintain subjects. In contrast, InstructP2P~\cite{instructpix2pix} achieves editing consistency by adding different numbers of input channels for models. These structures are tailored for specific tasks and lack generalizability. To this end, as shown in Fig.~\ref{fig:overall} (a), we concatenate VLM features with noisy latent and input them into a DIT Block for joint multi-head self-attention operations. After that,
the VLM features and noisy latent are processed by the FeedForward modules.
This allows the model to autonomously learn any level features (from overall consistency to subject consistency) for editing and generation. 
Notably, for FeedForward modules, we separate the VLM features and noisy latent, passing them through two distinct FeedForward modules with the same network structure.
Additionally, instead of using CLIP~\cite{clip} or T5~\cite{T5} as the text encoder, we introduce a Vision-Language Model (VLM) that enables joint understanding and encoding of both image and text prompts. For tasks that require high consistency, such as instruction-based editing and drag editing, we input the   VAE encoding of the source image into the DIT model to ensure strong consistency between the non-edited regions of the output image and the source image.

\begin{figure}[t]
    \begin{minipage}[b]{0.5\linewidth}
        \centering
        \includegraphics[width=1\columnwidth]{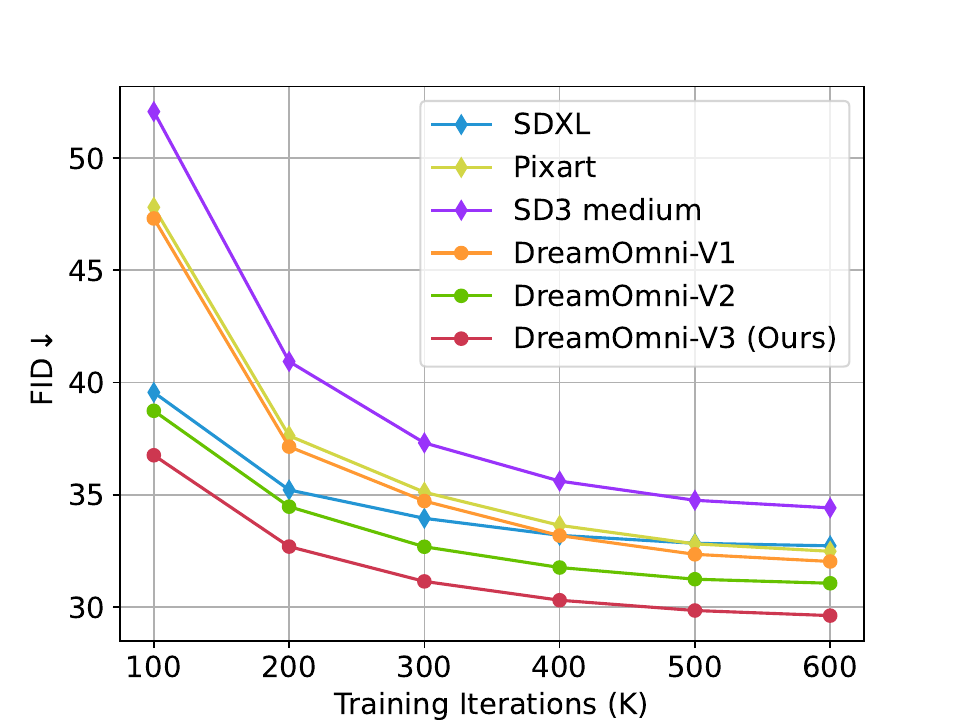} 
        \label{fig:fid_comparison}
    \end{minipage}
    \hfill
    \begin{minipage}[c]{0.45\linewidth}
    \vspace{-36mm}
        \centering
        \resizebox{1\linewidth}{!}{
        \begin{tabular}{l|cc}
            \toprule[0.2em]
            \textbf{Framework} & \textbf{Param (M)} & \textbf{Runtime (ms)} \\
            \midrule
            SDXL~\cite{sdxl}  & 838  & 28.59  \\
            Pixart~\cite{pixart} & 862  & 34.61  \\
            SD3~\cite{sd3}   & 896  & 25.73  \\
            \midrule
            DreamOmni-V1 & 828  & 28.32  \\
            DreamOmni-V2 & 836  & 25.14  \\
            DreamOmni-V3 (Ours) & 827  & 27.04  \\
            \bottomrule[0.2em]
        \end{tabular}
        }
        \label{table:pr_comparison}
    \end{minipage}
    \vspace{-6mm}
    \caption{ Comparison of different frameworks. The left figure shows the FID comparison among different frameworks, while the right table shows their number of parameters and runtime.}
    \label{fig:combined_comparison}
    \vspace{-2mm}
\end{figure}

In the current framework design, some works, such as DIT~\cite{DIT}, are compared under label-conditioned generation rather than T2I. However, T2I is inherently more complex than label-based generation, as it requires the integration and understanding of complex prompts. Additionally, many T2I models~\cite{sd3,LDM}, such as SDXL~\cite{sdxl}, are trained with different model sizes, datasets, and training settings. This variability makes it challenging to assess the impact of different model components on overall performance. Moreover, SDXL incorporates many Transformer blocks within its UNet structure. So, why does DIT outperform SDXL? To address this, we conducted extensive experiments as shown in Fig.~\ref{fig:combined_comparison}. We observe that DIT surpasses Unet because it allocates most of its computation to $2\times$ downsampled latent, whereas Unet allocates a greater proportion to $4\times$ downsampled latent. Since attention operations on $1\times$ latent cause memory burden, we further employ residual Conv blocks to refine generation details for $1\times$ latent.

Additionally, we observed that the use of long connections in the UNet framework can significantly accelerate the model's training convergence without compromising performance. As shown in Fig.~\ref{fig:overall} (a), we concatenate early and later features along the channel dimension and apply a linear layer to combine the two features. Notably, linear layers used for VLM features and noisy latent are different.

\subsection{Synthetic Data}
\label{sec:data}
In addition to the unified framework, we also require quite a few data to support joint training. While T2I data is readily available, creating and filtering accurate, high-quality data for tasks like instruction-based editing is far more challenging. To address this, we introduce a synthetic collage data pipeline that efficiently and accurately generates the required editing data. As shown in Fig.~\ref{fig:overall}~(b), our pipeline encompasses six tasks. Notably, this is not the full extent of our synthetic pipeline's capabilities; it is also capable of handling more complex task combinations.
 \begin{itemize}
 \item  \textit{T2I generation.} As shown in Fig.~\ref{fig:overall}~(b), in addition to the conventional T2I data, we further enhance the model’s performance on the T2I by incorporating synthetic data, specifically focusing on improving the text, shape, position, quantity, and color generation. Specifically, for the text, we randomly generate words or short phrases on a blank canvas using a variety of fonts, colors, thicknesses, and sizes. For the shape and quantity, we randomly create geometric shapes with different quantities, colors, and sizes, and arrange them on the canvas. Based on these attributes and their positions, we generate accurate prompts, which are then refined using a LLM. Furthermore, we use a diverse set of stickers and segmentation data for synthesis, placing them on the canvas and calculating their precise spatial relationships. These prompts are then created and further polished by the LLM to produce descriptions that are more natural.

\item  \textit{Inpainting \& Outpainting.} We randomly generate masks for smearing, blocks, and image edges. Notably, during training, in addition to feeding the masked image and its corresponding mask into the VLM for encoding, we also include the image caption with a $50\%$ probability.
 
\item  \textit{Instruction-based editing.} We categorize the task into three operations: addition, removal, and replacement. For both removal and replacement, we randomly select a background image and an object image to create the source image. In the removal case, the target image is simply the background image. For replacement, the target image is generated by replacing the object in the same location with a different object. Notably, for the addition, due to the need for the added object to be placed in a contextually appropriate position relative to the background, we use a blank background in the paper.

\item  \textit{Drag editing.} We categorize the data into three types: translation, scaling, and rotation. Notably, Instadrag~\cite{instadrag} treats each pair of drag points as a separate image, which is sparse and impractical due to the fixed number of drag points required. Therefore, we represent each drag point using the format $(x, y, dx, dy)$ as a prompt input, where $x$ and $y$ denote the coordinates of the drag points in the source image, and 
$dx$ and $dy$ represent the translation vector. Furthermore, we normalize these coordinates by dividing them by the image's width or height.

\item  \textit{Reference image generation.} We categorize the data into two types: image-conditioned generation, akin to ControlNet~\cite{controlnet}, and subject-driven generation.
For image-conditioned generation, we begin by selecting high-quality images and creating corresponding canny maps, depth maps, and segmentation masks as source images for training. 
For subject-driven generation, we synthesize source images and then randomly select objects within these images to create target images. The model is trained to generate new content based on specific attributes referenced from the source image, allowing for the flexible generation of varied scenes and subjects.

\item  \textit{Segmentation \& detection.} 
We randomly selected a background image and an object image. These images are then composited together to create a source image. Following this, we apply color manipulation or draw a bounding box around the object region based on the alpha channel of the object image, obtaining the target image.

\end{itemize}

Our synthetic collage data pipeline is efficient and precise, enabling the generation of billions of diverse images for large-scale pretraining and fine-tuning of DreamOmni.

\subsection{Model Training}
\label{sec:training}
After careful consideration, we chose a $2.5B$ parameter size for DreamOmni's DIT model. This size strikes a good balance between being user-friendly and ensuring powerful performance. For the VLM encoder, we directly adopt the Qwen2-VL~\cite{qwen2vl} $7B$ model, based on three reasons: \textbf{(1)} It supports image inputs of arbitrary resolution, \textbf{(2)} It delivers strong model performance, and \textbf{(3)} It is released under a permissive open-source license. The VLM features are derived from the penultimate layer of Qwen2-VL. Additionally, we use the FLUX-schnell's VAE as DreamOmni's VAE, which retains more latent channels, enabling the model to capture finer image details. Furthermore, we optimize DreamOmni using Rectified Flow~\cite{liu2022flow}, which performs the forward process by linearly interpolating between noise and data along a straight trajectory. we train DreamOmni using loss $\mathcal{L}$:
\vspace{-2mm}
\begin{equation}
\vspace{-2mm}
\mathcal{L}=\mathbb{E}\left(\left\|(\mathbf{z}-\mathbf{\epsilon})-v_\theta\left(\mathbf{z}_{t}, \mathbf{c}, t\right)\right\|_2^2\right),
\label{eq:sd_loss}
\end{equation}
where $\mathbf{z}_t=t \mathbf{z}+(1-t)\mathbf{\epsilon}$ represents the noised feature map at timestep $t$. Ground truth images are encoded by VAE into latent space to derive $\mathbf{z}$. Here, $\mathbf{\epsilon} \in \mathcal{N}(0, \mathbf{I})$ represents Gaussian noise, and $v_\theta$ refers to the DIT model. $\mathbf{c}$ indicates the conditional information.

The T2I training dataset consists of $125M$ images, including both the LAION dataset ($103M$)~\cite{laion5b} and our collected data ($22M$). These images have been annotated using InternVL2~\cite{internvl}. For synthetic data, such as T2I, instruction editing, inpainting \& outpainting, drag editing, and reference image generation, we generated $12M$ images respectively, bringing the total to approximately $60M$ images. Additionally, for segmentation \& detection, we created $8M$ images. Prior to training, to improve training speed usage, we first encoded prompts into VLM features.

\begin{table}[t]
  \centering
  \caption{Comparisons on GenEval~\cite{geneval}. Our model 
outperforms all current open models (including SOTA SD3-Medium~\cite{sd3}). }
\vspace{-2mm}
   \resizebox{1\linewidth}{!}{
    \begin{tabular}{l|ccccccc}
    \toprule[0.2em]
          &       & \multicolumn{2}{c}{\textbf{Objects}} &       &       &       &  \\
\cmidrule{3-4}    \textbf{Model} & \textbf{Overall} & \textbf{Single} & \textbf{Two}   & \textbf{Counting} & \textbf{Colors} & \textbf{Position} & \textbf{\shortstack{Color \\Attribution}  } \\
    \midrule
    SD 1.5~\cite{LDM} & 0.43  & 0.97  & 0.38  & 0.35  & 0.76  & 0.04  & 0.06 \\
    PixArt-alpha~\cite{pixart} & 0.48  & 0.98  & 0.5   & 0.44  & 0.8   & 0.08  & 0.07 \\
    SDv2.1 & 0.5   & 0.98  & 0.51  & 0.44  & 0.85  & 0.07  & 0.17 \\
    SDXL~\cite{sdxl}  & 0.55  & 0.98  & 0.74  & 0.39  & 0.85  & 0.15  & 0.23 \\
    SD-cascade &    0.52 &	0.98 &	0.57 &	0.47 &	0.87 &	0.08 &	0.15 \\
    SD3-Medium~\cite{sd3} &  \textbf{0.70} &	\textbf{0.99}	& \textbf{0.84} &	0.63 &	\textbf{0.88} &	0.28 &	\textbf{0.55} \\
    \midrule
    DreamOmni (Ours) &  \textbf{0.70} &	\textbf{0.99} &	0.81 &	\textbf{0.65} &	\textbf{0.88} &	\textbf{0.34} &	 0.54 \\
    \bottomrule[0.2em]
    \end{tabular}%
    }
  \label{tab:t2i}%
  \vspace{-2mm}
\end{table}%

\begin{figure*}[t]
	\centering
 \resizebox{1\linewidth}{!}{
	\includegraphics[height=4cm]{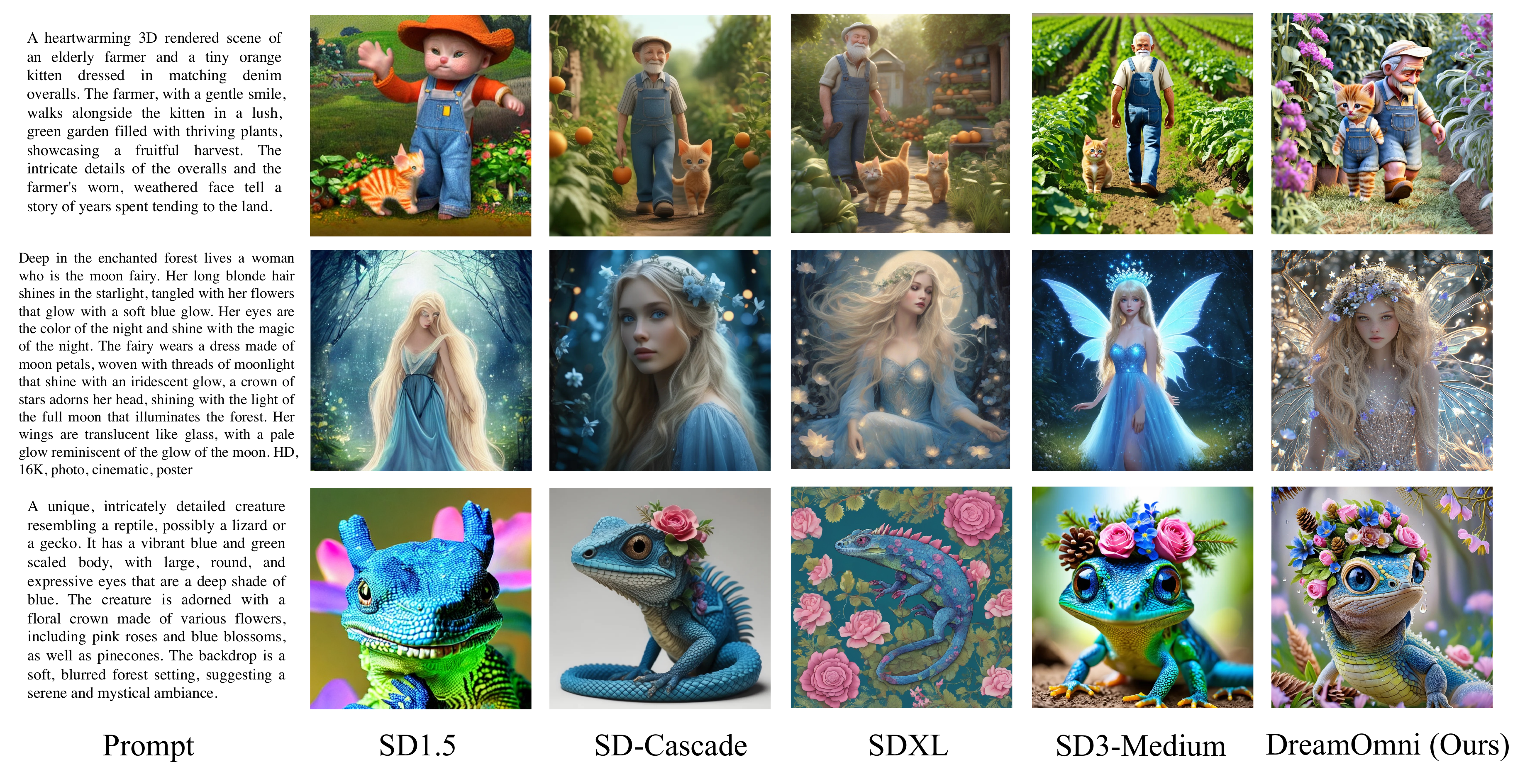}
 }
  \vspace{-7mm}
	\caption{
Visual comparison on \textbf{T2I generation}. Compared to other competitive methods (including SD3-Medium~\cite{sd3}, SDXL~\cite{sdxl}, SD-Cascade, and SD1.5~\cite{LDM}), our DreamOmni not only better adheres to user prompts but also generate more visually appealing results with delicate details, elegant composition, and so on.  }
	\label{fig:t2i}
\end{figure*}

\begin{figure*}[t]
	\centering
 \resizebox{1\linewidth}{!}{
	\includegraphics[height=4cm]{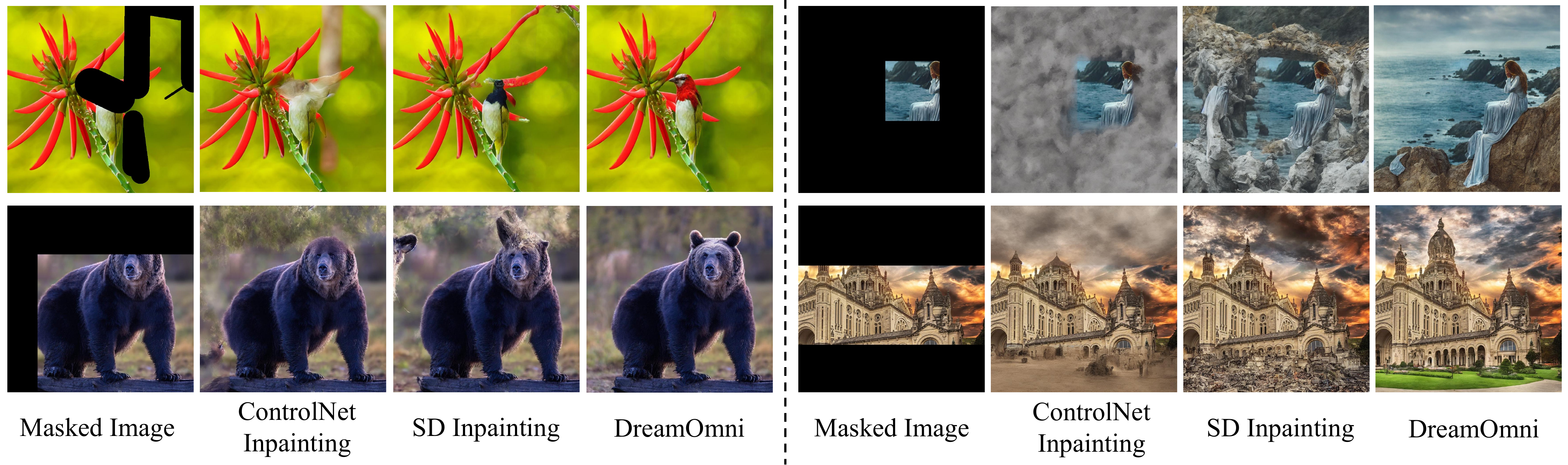}
 }
  \vspace{-6mm}
	\caption{Visual comparison on \textbf{inpainting $\&$ outpainting} between DreamOmni, ControlNet-Inpainting~\cite{controlnet} and SD-inpainting~\cite{LDM}.  }
	\label{fig:inpainting}
 \vspace{-2mm}
\end{figure*}

Our training process is divided into three stages. In the first stage, we train on images with $256\times256$ size, a batch size of $2048$, a learning rate of $1\times10^{-4}$, and $377K$ iterations. In the second stage, we train on images with $512\times512$ size, a batch size of $1024$, a learning rate of $5\times10^{-5}$, and $189K$ iterations. In the final stage, we select 12M high-quality T2I data and randomly sample 1M high-quality images from each type of synthetic data for training. We train models with $1024\times1024$ size, a batch size of $256$, a learning rate of $2\times10^{-5}$, and $140K$ iterations. All experiments are conducted on $64$ A100 GPUs. 
Additionally, to enable the model to generate images at varying resolutions, similar to the approach used in SDXL~\cite{sdxl}, we divide the images into $31$ buckets based on their aspect ratios, ranging from $4:1$ to $1:4$, during training.

\vspace{-1mm}
\section{Experiments}

\begin{figure*}[t]
	\centering
 \resizebox{1\linewidth}{!}{
	\includegraphics[height=4cm]{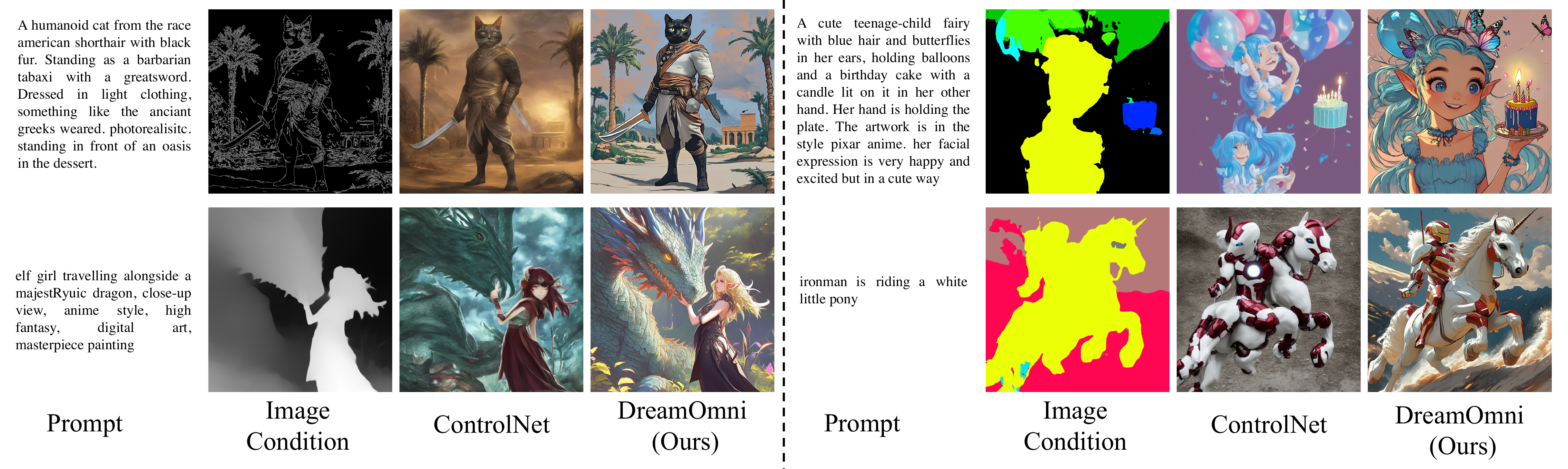}
 }
  \vspace{-6mm}
	\caption{Visual comparison on \textbf{image-conditioned generation}. Our DreamOmni excels at using canny, depth, and segmentation maps as conditions for image generation. Compared to the classic ControlNet~\cite{controlnet}, DreamOmni not only follows the user's prompts and image conditions more accurately but also generates content and color schemes that are visually more pleasing.}
	\label{fig:controlnet}
 \vspace{-3mm}
\end{figure*}

\begin{figure*}[t]
	\centering
 \resizebox{1\linewidth}{!}{
	\includegraphics[height=4cm]{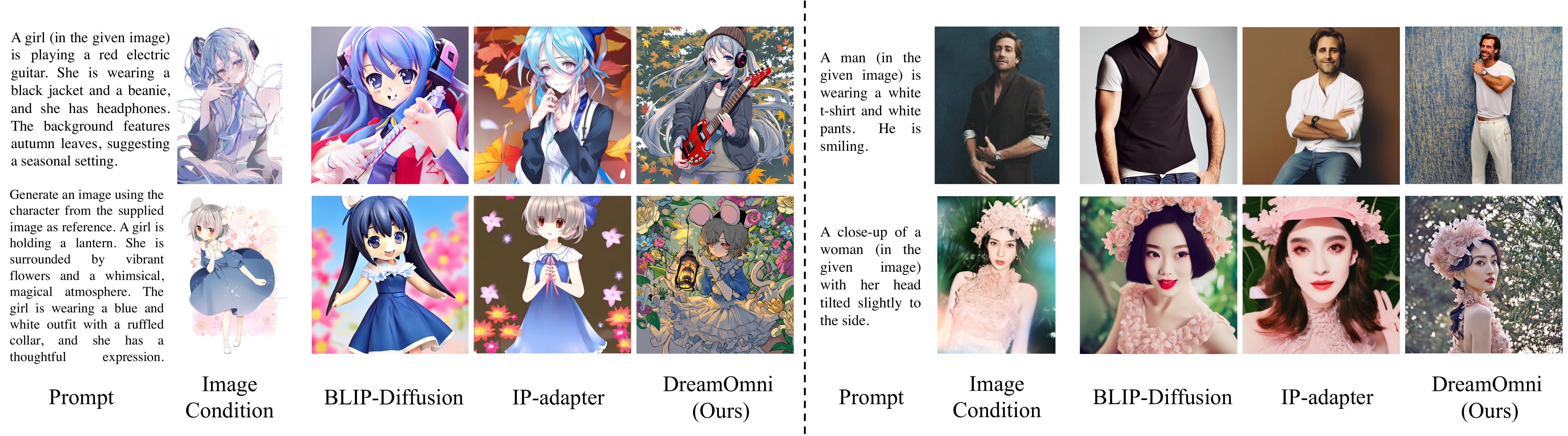}
 }
  \vspace{-6mm}
	\caption{Visual comparison on \textbf{subject-driven generation}. Compared with BLIP-Diffusion~\cite{blipdiffusion} and IP-adapter~\cite{ip-adapter}, DreamOmni excels at accurately following user prompts while preserving the specified subject. Furthermore, its success across both photo and anime cases underscores DreamOmni's generalization capabilities.  }
	\label{fig:subject}
 \vspace{-3mm}
\end{figure*}

\noindent\textbf{Evaluation on Framework.} 
We compared several T2I model frameworks under similar settings to identify effective components. Using the same VAE, CLIP text encoder, parameters, runtime, and LAION training/testing datasets, we evaluated the Unet-based SDXL~\cite{sdxl}, DIT-based Pixart~\cite{pixart}, SD3-Medium~\cite{sd3}, and our developed DreamOmni variations. Notably, as shown in Fig.~\ref{fig:combined_comparison}, to facilitate comparison, we do not use the full $2.5B$ parameters of DreamOmni, but instead adjust the parameters of all models to $0.85B$.
DreamOmni-V1 has two downsampling layers (2$\times$ and 4$\times$) like SDXL but lacks an Unet connection. DreamOmni-V2 builds on DreamOmni-V1 by adding the Unet connection, while DreamOmni-V3 takes DreamOmni-V2 further by focusing all DIT operations on 2$\times$ downsampled latent.
\textbf{(1)} Models with the Unet connection (SDXL, DreamOmni-V2, DreamOmni-V3) showed significantly faster convergence than those without (SD3-Medium, DreamOmni-V1). DreamOmni-V3, notably, converges four times faster than the SD3-Medium,  enhancing both training and fine-tuning efficiency.
\textbf{(2)} Comparing DreamOmni-V3 and DreamOmni-V2, we found that concentrating DIT block computations on higher-resolution latent (2$\times$) is more cost-effective.

\noindent\textbf{Evaluation on T2I generation.} As shown in Tab.~\ref{tab:t2i}, our synthetic data significantly enhances DreamOmni’s T2I generation capabilities in aspects such as quantity, color, and position, enabling our model to achieve SOTA results on GenEval~\cite{geneval}. Notably, SD3-Medium~\cite{sd3} is a $2B$ open-source SOTA T2I model, which has a similar parameters as our DreamOmni. Furthermore, qualitative results are shown in Fig.~\ref{fig:t2i}. We can see that DreamOmni’s outputs are not only more visually pleasing but also align more accurately with the given prompts.

\noindent\textbf{Evaluation on Inpainting.} We compare our DreamOmni with ControlNet-Inpainting~\cite{controlnet} and SD-Inpainting~\cite{LDM}  on our high-quality evaluation datasets to evaluate its performance. The quantitative results, presented in Tab.~\ref{tab:inpainting}
, demonstrate that DreamOmni significantly outperforms both ControlNet-Inpainting and SD-Inpainting, highlighting its superior generation quality and coherence. Visual results (Fig.~\ref{fig:inpainting}) further emphasize that DreamOmni excels in generating fine details, surpassing both ControlNet-Inpainting and SD-Inpainting. Additionally, DreamOmni effectively handles large mask holes, generating realistic content instead of blurry and incoherent outputs.

\noindent\textbf{Evaluation on Reference Image Generation.} For image-conditioned generation, we conducted a comparison with ControlNet~\cite{controlnet} on canny, depth, and segmentation image conditions. Visual results are presented in Fig.~\ref{fig:controlnet}. These results reveal that, across all tested conditions, our DreamOmni significantly outperforms ControlNet. Our method not only adheres more faithfully to both the image condition and the prompt but also demonstrates enhanced visual quality with better composition and richer detail.

For subject-driven image generation, we compared our method with competitive approaches, including BLIP-Diffusion~\cite{blipdiffusion} and IP-Adapter~\cite{ip-adapter}. To demonstrate DreamOmni's powerful generalization, we validated their performance on both anime and photographic images. Visual results are shown in Fig.~\ref{fig:subject}. Compared to other methods, our DreamOmni not only preserves the specified subject effectively but also follows the prompt well.

\begin{table}[t]
  \centering
  \caption{Quantitative comparison for inpainting $\&$ outpainting.} 
  \vspace{-2mm}
  \resizebox{1\linewidth}{!}{
    \begin{tabular}{l|rr|rr}
    \toprule[0.2em]
     & \multicolumn{2}{c|}{\textbf{Inpainting}} & \multicolumn{2}{c}{\textbf{Outpainting}} \\
\cmidrule{2-5}     \textbf{Model}     & \multicolumn{1}{c}{\textbf{FID$\downarrow$}} & \multicolumn{1}{c|}{\textbf{LPIPS$\downarrow$}} & \multicolumn{1}{c}{\textbf{FID$\downarrow$}} & \multicolumn{1}{c}{\textbf{LPIPS$\downarrow$}} \\
    \midrule
    SD-inpainting~\cite{LDM} & 1.3522 & 0.1560 & 2.9179 & 0.2475 \\
    ControlNet-inpainting~\cite{controlnet} & 1.8393 & 0.1594 & 4.2337 & 0.2521 \\
    \midrule
    DreamOmni (Ours) & \textbf{0.8371} & \textbf{0.1203} & \textbf{1.6926} & \textbf{0.1995} \\
    \bottomrule[0.2em]
    \end{tabular}%
    }
  \label{tab:inpainting}%
  \vspace{-3mm}
\end{table}%

\begin{figure*}[t]
	\centering
 \resizebox{1\linewidth}{!}{
	\includegraphics[height=4cm]{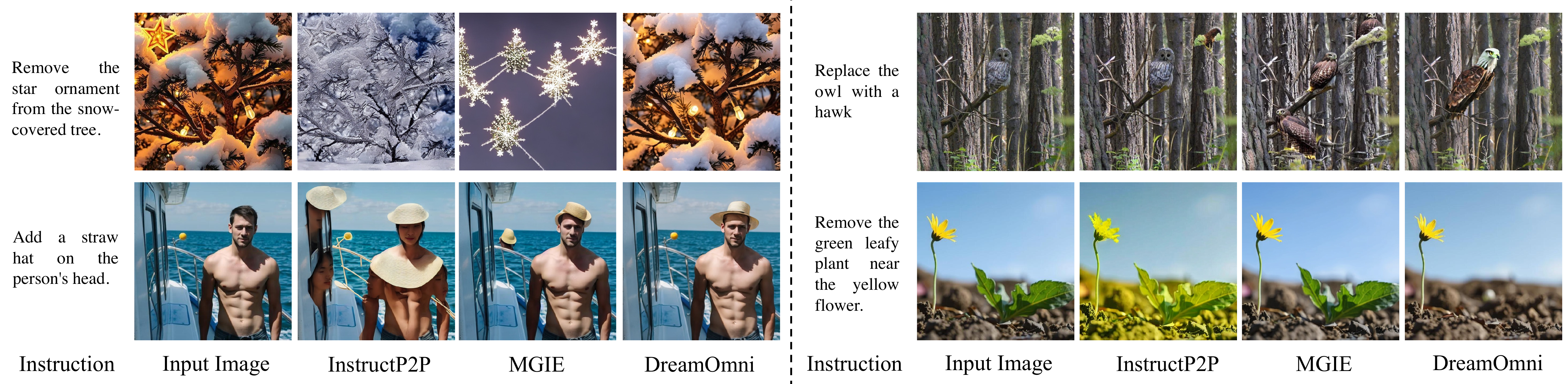}
 }
  \vspace{-5mm}
	\caption{Visual comparison on \textbf{instruction-based editing}. Our DreamOmni achieves more precise editing (including addition, removal, and replacement) compared to competitive methods such as MGIE~\cite{mgie} and InstructP2P~\cite{instructpix2pix}.  }
	\label{fig:ip2p}
 \vspace{-2mm}
\end{figure*}

\noindent\textbf{Evaluation on Instruction Editing.}
We compare DreamOmni with competitive methods such as MGIE~\cite{mgie} and InstructP2P~\cite{instructpix2pix}. The visual results are presented in Fig.~\ref{fig:ip2p}. We can see that DreamOmni performs more accurate edits, including addition, removal, and replacement. Specifically, our editing results exhibit superior consistency with minimal changes to non-edited areas and higher-quality generation of the edited content. This further validates that our synthetic data pipeline is an efficient and effective method for creating instruction-based editing datasets, enabling models to learn precise instruction-based editing. Moreover, the efficiency of our synthetic data pipeline allows for models to easily scale up diverse training data.

\begin{figure}[t]
	\centering
 \resizebox{1\linewidth}{!}{
	\includegraphics[height=4cm]{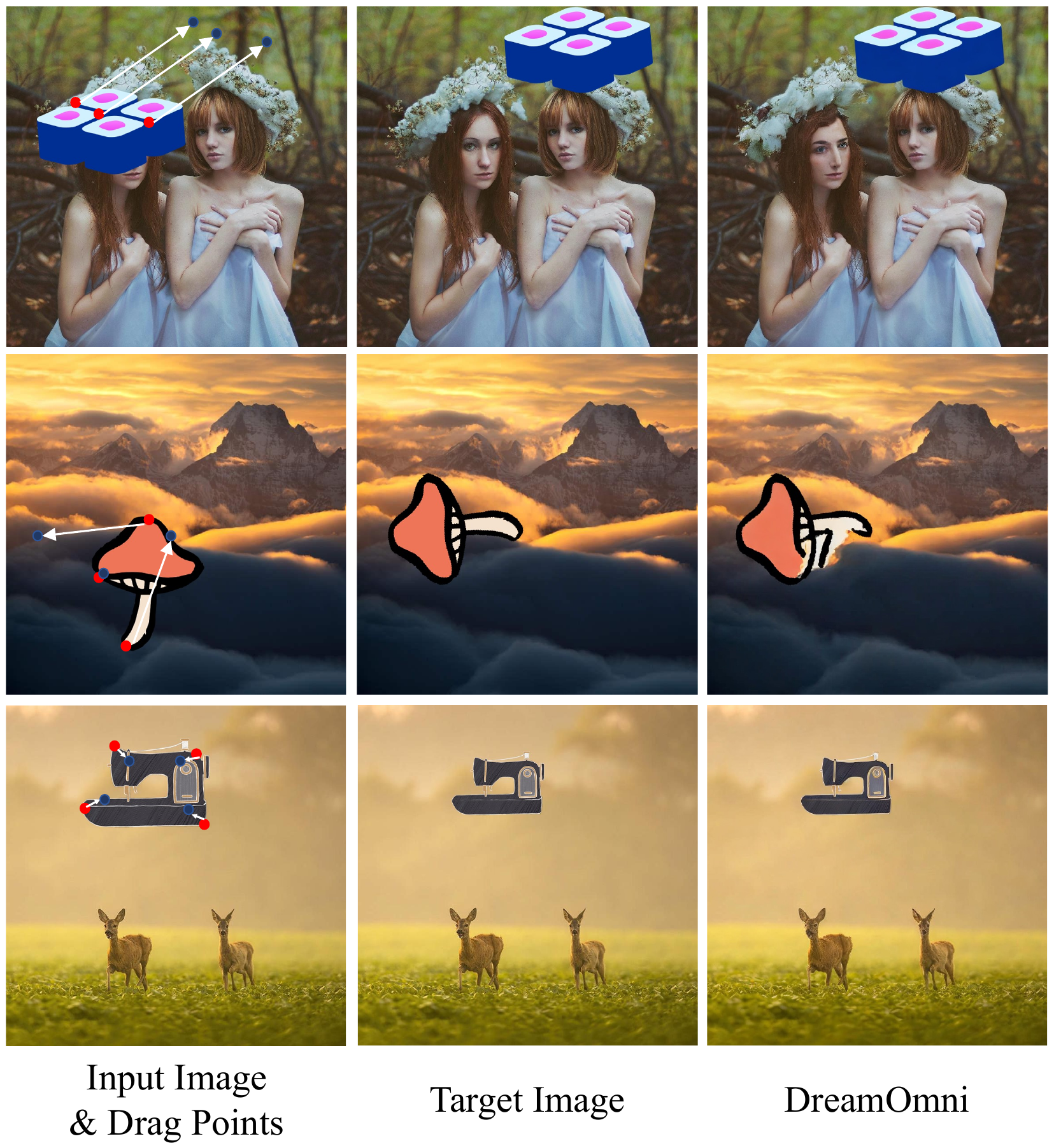}
 }
  \vspace{-6mm}
	\caption{Visual results of \textbf{drag editing}: DreamOmni accurately performs translation, rotation, and scaling edits.    }
	\label{fig:drag}
 \vspace{-3mm}
\end{figure}

\noindent\textbf{Evaluation on Drag Editing.} We evaluate DreamOmni on our synthetic evaluation dataset. The visual results are shown in Fig.~\ref{fig:drag}. \textbf{(1)} Compared to the target image, we can see that DreamOmni can accurately perform translation, rotation, and scaling drag edits. \textbf{(2)} for translation and scaling, DreamOmni can maintain the integrity of the dragged object. However, large-scale rotation operations are more challenging for DreamOmni, as they involve complex transformations of the object itself, which may lead to deformations of the edited object. \textbf{(3)} The results demonstrate the effectiveness of our synthetic data pipeline for drag editing, and encoding the drag point positions and displacement information as instruction inputs allows the model to learn precise drag edits (shown in Fig.~\ref{fig:overall}).
\section{Conclusion}
Current T2I foundation models lack a unified framework and training for downstream tasks, such as image editing. To address this, we introduce DreamOmni, a unified model for T2I generation and editing. We evaluate the frameworks of existing models under fair settings and consider the specific needs of various editing tasks. From this analysis, we develop a framework that integrates T2I with various editing tasks. Besides,
a challenge in training editing models is the creation of high-quality, large-scale editing data, which is inefficient. To overcome this, we designed a synthetic collage data pipeline capable of efficiently generating quite a few precise, high-quality editing data. Moreover, the pipeline enhances the model's generation accuracy in text, position, quantity, color, and geometry.
By jointly training on T2I and multi-task synthetic data, we develop a native, unified model for both image generation and editing. T2I training strengthens the model’s grasp of specific concepts and improves generation quality, while editing training enables it to handle the requirements of editing tasks.

{
    \small
    \bibliographystyle{ieeenat_fullname}
    \bibliography{main}
}


\end{document}